\documentclass[runningheads]{llncs}
\usepackage[T1]{fontenc}
\usepackage{booktabs}
\usepackage{amsmath}
\usepackage{graphicx}
\usepackage{color}
\usepackage[hidelinks]{hyperref}
\begin{document}
\title{A Survey of Recent DNN Architectures on the TIMIT Phone Recognition Task
\thanks{This paper was supported by the project no. P103/12/G084 of the Grant Agency of the Czech Republic and by the grant of the University of West Bohemia, project No. SGS-2016-039.
Access to computing and storage facilities owned by parties and projects contributing to the National Grid Infrastructure MetaCentrum provided under the programme "Projects of Large Research, Development, and Innovations Infrastructures" (CESNET LM2015042), is greatly appreciated.}}
%
%
\author{Josef Mich\'{a}lek\orcidID{0000-0001-7757-3163} \and
Jan Van\v{e}k\orcidID{0000-0002-2639-6731}}
\authorrunning{J. Mich\'{a}lek, J. Van\v{e}k}
%
\institute{University of West Bohemia \\ Univerzitn\'{i} 8, 301 00 Pilsen, Czech Republic\\
\email{\{orcus,vanekyj\}@kky.zcu.cz}}
\maketitle              
\begin{abstract}
In this survey paper, we have evaluated several recent deep neural network (DNN) architectures on a TIMIT phone recognition task. We chose the TIMIT corpus due to its popularity and broad availability in the community. It also simulates a low-resource scenario that is helpful in minor languages. Also, we prefer the phone recognition task because it is much more sensitive to an acoustic model quality than a large vocabulary continuous speech recognition (LVCSR) task.
In recent years, many DNN published papers reported results on TIMIT. However, the reported phone error rates (PERs) were often much higher than a PER of a simple feed-forward (FF) DNN. That was the main motivation of this paper: To provide a baseline DNNs with open-source scripts to easily replicate the baseline results for future papers with lowest possible PERs. According to our knowledge, the best-achieved PER of this survey is better than the best-published PER to date.
\end{abstract}
\section{Introduction}
The Texas Instruments/Massachusetts Institute of Technology (TIMIT) corpus of read speech \cite{TIMIT} is available since 1993 in LDC as an LDC93S1 corpus. It has been designed for the development and evaluation of automatic speech recognition systems. TIMIT contains speech from 630 speakers representing 8 major dialect divisions of American English, each speaking 10 phonetically-rich sentences. The TIMIT corpus includes time-aligned orthographic, phonetic, and word transcriptions, as well as speech waveform data for each spoken sentence. Very valuable part is a definition of training, development, and test sets. It helped to develop the TIMIT corpus to be a very popular phone recognition benchmark task.

Very detailed overview until 2011 was published by Lopes and Fernando in \cite{Lopes2011}. It mapped the pre-DNN era and the start of the deep learning represented by a Mohamed at al. monophone deep belief network (DBN) \cite{Mohammed_TASL_2011} with PER 20.7\% on the core test set. A triphone version of the DBN with a speaker adaptive training and a fMLLR adaptation was developed by Bagher BabaAli and Karel Vesely in the TIMIT Kaldi example s5 \cite{Kaldi_GIT}. The Kaldi example achieved PER 18.5\% on the core test set. Better results were then obtained by DNNs with rectified linear units (ReLU). The ReLU DNNs do not need the DBN pretraining and, if dropout is applied, they perform well on held out data. Laszlo Toth reported PER 17.76\% on core test set with a convolutional bottle neck ReLU DNN in \cite{Toth_IS2013} and a year later he reported PER 16.5\% with a 2D convolutional bottleneck maxout DNN in \cite{Toth_IS2014}.
We also reported PER 16.5\% with an ensemble of DBN DNNs augmented by regularization post-layer \cite{Vanek_2017}. Taesup Moon then achieved stable PER 16.9\% with a droupout bi-directional long-short term memory recurrent DNN (DBLSTM) and a peak PER with a larger net up to 16.29\% \cite{Moon2015}.

In this paper, we have evaluated several recent deep neural network architectures. We also published our scripts to easily repeat our work and results. We followed the Kaldi s5 example and limit the experiments to a triphone model obtained by the Kaldi example together with the fMLLR speaker adapted training, development, and test data. First, we evaluated a simple feed-forward ReLU DNNs. Then, we tested time delay neural networks (TDNNs). Finally, we evaluated long-short term memory (LSTM) recurrent DNNs which gave the lowest PER. Because of the common feature processing stage, we did not try any 2D convolutional DNNs. We plan to investigate them as a future work. 

\section{Neural Network Architectures}
\subsection{Feed Forward DNNs}
First DNNs used a sigmoid activation function, which suffers from the vanishing gradient problem.
Hinton et al. in \cite{hinton2006fast} proposed a greedy layer-wise unsupervised pre-training learning procedure.
This procedure relies on the training algorithm of restricted Boltzmann machines (RBM) and initializes the
parameters of a deep belief network (DBN), a generative model with many layers
of hidden causal variables. The greedy layer-wise unsupervised training strategy
helps the optimization by initializing weights in a region near a good local minimum, but also implicitly acts as a sort of regularization that brings better generalization and encourages internal distributed representations that are high-level abstractions of the input \cite{larochelle2009exploring}.

Later DNNs used ReLU, that do not suffer from the vanishing gradient problem.
Therefore, pre-training is not necessary.
On the other hand, the ReLU DNNs are more prone to overfitting.
The most effective regularization technique is dropout \cite{srivastava2014dropout}.

\subsection{Time Delay Neural Network}
The time delay neural network (TDNN) is a network designed to classify patterns shift-invariantly.
It was first proposed to classify phonemes in speech recognition systems \cite{waibel1990phoneme}.

In standard DNN, initial layers learn an affine transform of the entire temporal context.
However, in TDNN, the initial transforms are learnt on narrow contexts and the deeper layers process the hidden activations from a wider temporal context.
Therefore, the higher layers have the ability to learn wider temporal dependencies.
Usually, each layer operates on different temporal resolution, which increases as we go higher in the layers.
The transforms in TDNN are tied across time steps and this is why TDNNs are seen as precursors to convolutional networks.

The hyperparameters defining a TDNN network are input context size for each layer and a number of filters in each layer.

In our work, we used ReLU as an activation function for TDNN.
Other authors such as Peddinti use \emph{p}-norm nonlinearity, although in \cite{peddinti2015time} he proposes switching to ReLU due to the better results of his preliminary experiments.

\subsection{Long Short-Term Memory}
\label{sec:lstm}
Long short-term memory (LSTM) is a widely used type of recurrent neural network (RNN).
Standard RNNs suffer from both exploding and vanishing gradient problems.

The exploding gradient problem can be solved simply by truncating the gradient.
On the other hand, the vanishing gradient problem is harder to overcome.
It does not simply cause the gradient to be small; the gradient components corresponding to long-term dependencies are small while the components corresponding to short-term dependencies are large.

The LSTM was proposed in 1997 by Hochreiter and Scmidhuber \cite{hochreiter1997long} as a solution to the vanishing gradient problem.
Let $c_t$ denote a hidden state of a LSTM.
The main idea is that instead of computing $c_t$ directly from $c_{t-1}$ with matrix-vector product followed by an activation function, the LSTM computes $\Delta c_t$ and adds it to $c_{t-1}$ to get $c_t$.
The addition operation is what eliminates the vanishing gradient problem.

Each LSTM cell is composed of smaller units called gates, which control the flow of information through the cell.
The forget gate controls what information will be discarded from the cell state, input gate controls what new information will be stored in the cell state and output gate controls what information from the cell state will be used in the output.

The LSTM has two hidden states, $c_t$ and $h_t$.
The state $c_t$ fights the gradient vanishing problem while $h_t$ allows the network to make complex decisions over short periods of time.
There are several slightly different LSTM variants.
The architecture used in this paper is specified by the following equations:
\begin{align*}
i_t &= \sigma(W_xi x_t + W_hi h_{t-1} + b_i) \\
f_t &= \sigma(W_xf x_t + W_hf h_{t-1} + b_f) \\
o_t &= \sigma(W_xo x_t + W_ho h_{t-1} + b_o) \\
c_t &= f_t \ast c_{t-1} + i_t \ast \tanh(W_xc x_t + W_hc h_{t-1} + b_c) \\
h_t &= o_t \ast \tanh(c_t)
\end{align*}

\section{Experiments}

The TIMIT corpus contains recordings of phonetically-balanced prompted English speech. It was recorded using a Sennheiser close-talking microphone at 16 kHz rate with 16 bit sample resolution. TIMIT contains a total of 6300 sentences (5.4 hours), consisting of 10 sentences spoken by each of 630 speakers from 8 major dialect regions of the United States. All sentences were manually segmented at the phone level.

The prompts for the 6300 utterances consist of 2 dialect sentences (SA), 450 phonetically compact sentences (SX) and 1890 phonetically-diverse sentences (SI).

The training set contains 3696 utterances from 462 speakers. The core test set consists of 192 utterances, 8 from each of 24 speakers (2 males and 1 female from each dialect region). The training and test sets do not overlap. 
\subsection{Speech Data, Processing, and Test Description}
As mentioned above, we used TIMIT data available from LDC as a corpus LDC93S1. Then, we ran the Kaldi TIMIT example script s5, which trained various NN-based phone recognition systems with a common HMM-GMM tied-triphone model and alignments. The common baseline system consisted of the following methods: It started from MFCC features which were augmented by $\Delta$ and $\Delta\Delta$ coefficients and then processed by LDA. Final feature vector dimension was 40. We obtained final alignments by HMM-GMM tied-triphone model with 1909 tied-states (may vary slightly if rerun the script). We trained the model with MLLT and SAT methods, and we used fMLLR for the SAT training and a test phase adaptation. We dumped all training, development and test fMLLR processed data, and alignments to disk. Therefore, it was easy to do compatible experiments from the same common starting point. We employed a bigram language/phone model for final phone recognition. A bigram model is a very weak model for phone recognition; however, it forced focus to the acoustic part of the system, and it boosted benchmark sensitivity. The training, as well as the recognition, was done for 48 phones. We mapped the final results on TIMIT core test set to 39 phones (as it is usual by processing TIMIT corpus), and phone error rate (PER) was evaluated by the provided NIST script to be compatible with previously published works. In contrast to the Kaldi recipe, we used a different phone decoder. It is a standard Viterbi-based triphone decoder. It gives better results than the Kaldi standard WFST decoder on the TIMIT phone recognition task.
We have used an open-source Chainer 3.0 DNNs Python tranining tool that supports NVidia GPUs \cite{Chainer}. It is multiplatform and easy to use. 

\subsection{Feed-Forward DNNs}
First, we re-trained feed-forward (FF) DNN with sigmoid activation function from the Kaldi example. We used the identical topology and the DBN pre-trained parameters.

In the other experiments with FF DNNs, we used a simple DNNs with ReLU without any pre-training. We used lower dropout $p=0.2$, we have obtained better results than $p=0.5$ like in \cite{Toth_IS2013}. We stacked 11 input fMLLR feature frames to 440 NN input dimension, like in Kaldi example s5. All the input vectors were transformed by an affine transform to normalize input distribution. We have tested a range from 6 to 9 hidden layers with 512, 1024, and 2048 ReLU neurons. The final softmax layer had 1909 neurons. We used SGD with momentum 0.9. The learning rate was three-times reduced according to development data training criterion change. Together with the learning rate reduction, the batch size was gradually increased from initial 256 to 512, 1024, and final 2048.

Besides ReLU, we tried also other activation functions. Leaky-ReLU gave almost identical error rates and criterion like ReLU. We have tested also maxout units (1/2, 1/4, and 1/8) that gave worse results than ReLU. 

\subsection{Time Delay Neural Network}
First we used a network with 4 layers and context sizes 5,5,9,9 as in \cite{peddinti2015time} without sub-sampling.
We used dropout $p=0.2$.
Input data were stacked to the size required by the context sizes and normalized.
We also evaluated context sizes 5,5,5,5 and 9,9,9,9.
We tested hidden layers with 256, 512 and 1024 filters with ReLU activation function.
The final layer was again softmax with 1909 neurons.

The networks were trained first using Adam optimization algorithm and then using SGD with momentum 0.9.
SGD training stage was used three times, each with lower learning rate.
Batch size was the same as in FF DNN case.

\subsection{Long Short-Term Memory}
We used standard LSTM architecture as specified in section \ref{sec:lstm}.
We tried the number of hidden layers in range from 2 to 6 and 256, 512 and 1024 LSTM cells in each hidden layer.
Input data were transformed to normalize the input distribution.
We used output time delay equal to 5 time steps.
Dropout used was again $p=0.2$.

The network was trained first using Adam and then momentum SGD as in TDNN case.
Batch size used was 512 for Adam and 128 for SGD training stages.

\subsection{Results}
\begin{table}[t]
    \center
    \caption{Feed-Forward DNN Phone Error Rate}
    \label{tbl:ff}
    \setlength{\tabcolsep}{2mm}
    \begin{tabular}{l|ccc|ccc}
        \toprule
        & \multicolumn{3}{c|}{Development PER} & \multicolumn{3}{c}{Test PER} \\
        \multicolumn{1}{c|}{Network} & Min & Max & Avg & Min & Max & Avg \\
        \midrule
        FF sigmoid 7x1024 & 16.17 & 16.48 & 16.31 & 16.76 & 17.24 & 17.04 \\
        FF ReLU 6x512 & 15.97 & 16.64 & 16.40 & 17.34 & 18.03 & 17.63 \\
        FF ReLU 6x1024 & 16.06 & 16.37 & 16.23 & 16.90 & 17.34 & 17.09 \\
        FF ReLU 6x2048 & 15.94 & 16.48 & 16.27 & 16.88 & 17.39 & 17.17 \\
        FF ReLU 7x512 & 16.05 & 16.49 & 16.26 & 17.20 & 18.09 & 17.50 \\
        FF ReLU 7x1024 & 15.70 & 16.26 & 15.95 & 16.62 & 17.23 & 16.93 \\
        FF ReLU 7x2048 & 15.92 & 16.49 & 16.17 & 16.83 & 17.41 & 17.06 \\
        FF ReLU 8x512 & 16.23 & 16.65 & 16.43 & 17.33 & 18.30 & 17.72 \\
        FF ReLU 8x1024 & 15.79 & 16.21 & 16.02 & 16.80 & 17.27 & 17.03 \\
        FF ReLU 8x2048 & 15.78 & 16.24 & 15.99 & 16.66 & 17.13 & 16.91 \\
        FF ReLU 9x512 & 16.34 & 16.57 & 16.50 & 17.38 & 17.71 & 17.57 \\
        FF ReLU 9x1024 & 15.73 & 16.01 & 15.91 & 16.49 & 17.41 & 16.99 \\
        FF ReLU 9x2048 & 15.61 & 16.08 & 15.91 & 16.69 & 17.23 & 16.96 \\
        \bottomrule
    \end{tabular}
\end{table}

\begin{table}[t]
    \footnotesize
    \center
    \caption{TDNN Phone Error Rate}
    \label{tbl:tdnn}
    \setlength{\tabcolsep}{2mm}
    \begin{tabular}{ll|ccc|ccc}
        \toprule
        & & \multicolumn{3}{c|}{Development PER} & \multicolumn{3}{c}{Test PER} \\
        \multicolumn{1}{c}{Context Size} & \multicolumn{1}{c|}{\# Filters} & Min & Max & Avg & Min & Max & Avg \\
        \midrule
        5,5,5,5 & 256 & 17.55 & 17.65 & 17.61 & 18.27 & 18.75 & 18.57 \\
        5,5,5,5 & 512 & 17.19 & 17.38 & 17.30 & 18.02 & 18.30 & 18.20 \\
        5,5,5,5 & 1024 & 17.33 & 17.72 & 17.46 & 17.77 & 18.66 & 18.23 \\
        5,5,9,9 & 256 & 17.71 & 17.96 & 17.82 & 18.49 & 18.84 & 18.67 \\
        5,5,9,9 & 512 & 17.32 & 17.59 & 17.50 & 18.30 & 18.82 & 18.52 \\
        5,5,9,9 & 1024 & 17.67 & 17.83 & 17.79 & 18.50 & 19.24 & 18.85 \\
        9,9,9,9 & 256 & 17.95 & 18.28 & 18.14 & 18.71 & 19.03 & 18.90 \\
        9,9,9,9 & 512 & 17.61 & 18.04 & 17.86 & 18.66 & 19.07 & 18.81 \\
        9,9,9,9 & 1024 & 18.26 & 20.34 & 18.91 & 18.82 & 21.07 & 19.72 \\
        \bottomrule
    \end{tabular}
\end{table}

After we trained all the networks, we have evaluated their performance on the development and test dataset.
The table \ref{tbl:ff} contains the minimum, maximum and average phone error rates for each type of used feed-forward networks.
The FF network with sigmoid activation function and DBN pre-training has average 17.04 \% PER, but it was outperformed by several simple FF networks with ReLU without pretraining.
The best average PER we have achieved is 16.91 \% PER in the case of 8x2048 FF network.
However, all the deeper networks with at least 1024 neurons in the hidden layers have similar performance.

The table \ref{tbl:tdnn} contains the results for TDNN networks.
All the results are worse than the results for our FF networks.
Interestingly enough, we obtained the best results for context sizes 5,5,5,5.
The larger context sizes resulted in worse network performance.
Also, the networks with higher number of filters have better performance.
The best average PER we have received was 18.20 \% for the network with the context size 5,5,5,5 and 512 filters in each layer.

The results for our LSTM experiments are in the table \ref{tbl:lstm}.
Our LSTM networks have better performance than other network types used in this work.
We have received the best average PER, 15.58 \%, for the network with 4 hidden layers each with 1024 LSTM units.
The best PER from all experiments was 15.02 \%, also belonging to the network with 4 hidden layers with 1024 LSTM units.
However, we have also trained other networks with similar PER to the best one.
The network with 3 layers with 1024 units and the network with 5 layers with 1024 units have the second and third best PER, 15.69 \% and 15.71 \% respectively.
The networks with 256 units in each layer have comparable or worse PER than FF networks.
The network with 6 layers with 1024 units has worse performance than less deep layers due to overfitting.

\begin{table}[t]
    \footnotesize
    \center
    \caption{LSTM Phone Error Rate}
    \label{tbl:lstm}
    \setlength{\tabcolsep}{2mm}
    \begin{tabular}{l|ccc|ccc}
        \toprule
        & \multicolumn{3}{c|}{Development PER} & \multicolumn{3}{c}{Test PER} \\
        \multicolumn{1}{c|}{Network} & Min & Max & Avg & Min & Max & Avg \\
        \midrule
        2x256 & 16.21 & 16.99 & 16.62 & 16.96 & 18.13 & 17.51 \\
        3x256 & 15.63 & 16.40 & 16.04 & 16.37 & 17.08 & 16.76 \\
        4x256 & 15.39 & 16.19 & 15.80 & 16.09 & 16.99 & 16.58 \\
        5x256 & 15.39 & 15.95 & 15.79 & 15.98 & 17.03 & 16.49 \\
        6x256 & 15.73 & 16.21 & 16.00 & 15.97 & 17.16 & 16.73 \\
        2x512 & 15.14 & 16.07 & 15.59 & 16.11 & 16.91 & 16.41 \\
        3x512 & 14.82 & 15.39 & 15.11 & 15.77 & 16.55 & 16.05 \\
        4x512 & 14.75 & 15.32 & 15.08 & 15.69 & 16.19 & 15.96 \\
        5x512 & 14.65 & 15.31 & 14.97 & 15.36 & 16.27 & 15.83 \\
        6x512 & 14.91 & 15.59 & 15.25 & 15.80 & 16.29 & 16.02 \\
        2x1024 & 14.92 & 15.36 & 15.08 & 15.62 & 16.17 & 15.97 \\
        3x1024 & 14.37 & 15.02 & 14.68 & 15.38 & 15.95 & 15.69 \\
        4x1024 & 14.43 & 15.16 & 14.67 & 15.02 & 15.84 & 15.58 \\
        5x1024 & 14.49 & 14.85 & 14.66 & 15.34 & 16.04 & 15.71 \\
        6x1024 & 14.82 & 15.20 & 15.04 & 15.34 & 16.48 & 15.87 \\
        \bottomrule
    \end{tabular}
    \vspace{-0.2cm}
\end{table}

\section{Conclusion}
We have trained several neural network types on the TIMIT phone recognition task.
Each of the networks was trained several times, 4 times in case of TDNN and 10 times for other network types.
We have evaluated the phone error rate (PER) for each trained model and showed the minimum, maximum and average PER in tables.

First we trained FF networks with ReLU and obtained 16.49 \% PER as our best result.
We have also tried FF network with sigmoid activation function and DBN pre-training, but the best model has worse PER than some of our FF ReLU networks, 16.76 \%.
Then we trained TDNN networks with several settings, but we couldn't get better results than our simple FF networks.
Our best TDNN network has 17.77 \% PER.
The last network type we trained was recurrent LSTM network.
We have found that the best performing networks have 4 layers each with 1024 LSTM units.
Their average PER was 15.58 \%, although networks with 3 or 5 layers have similar results.
The best result of all of our experiments was 15.02 \% PER for LSTM network with 4 layers each with 1024 units.
These results are the best results published to date according to our knowledge.

The scripts used in our experiments are freely available at\\ \url{https://github.com/OrcusCZ/NNAcousticModeling}.
%
%
%
\bibliographystyle{splncs04}
\bibliography{TSD2018_Michalek_Vanek}
\end{document}